\documentclass{article}

\usepackage{arxiv}
\usepackage{amsmath}
\usepackage{multirow}
\usepackage{natbib}
\usepackage[utf8]{inputenc} 
\usepackage[T1]{fontenc}    
\usepackage{hyperref}       
\usepackage{url}            
\usepackage{booktabs}       
\usepackage{amsfonts}       
\usepackage{nicefrac}       
\usepackage{microtype}      
\usepackage{lipsum}
\usepackage{graphicx}
\usepackage{hyperref}
\graphicspath{ {./images/} }

\title{Semantic Step Prediction:\\Multi-Step Latent Forecasting in LLM Reasoning Trajectories via Step Sampling}

\author{%
  Yuan Yidi \\
  Home Team Science and Technology Agency \\
  \texttt{yuan\_yidi@htx.gov.sg} \\
}

\begin{document}

\maketitle

\begin{abstract}
[This arXiv paper is entirely derived from work conducted as part of the NUS CS$5260$ course.] Semantic Tube Prediction (STP) leverages representation geometric to regularize LLM hidden-state trajectories toward locally linear geodesics during fine-tuning, thereby greatly improving data efficiency.  The original STP recipe samples random token sub-spans, which is compatible with the base large language model (LLM) training architecture. Inspired by STP, we are interested to investigate whether the sampling position can further enhance the semantic structure of multi-step reasoning, and hence affect its geometric impact. We applied STP at consecutive semantic reasoning step boundaries and achieved $168\times$ more accurate multi-step latent prediction than frozen baselines on ProcessBench (3,400 samples), compared to only $4\times$ for the random-token STP. Probing the latent manifold with a learned non-linear predictor reveals that STP-shaped trajectories are smooth curves, not straight lines: a 3-layer MLP reduces prediction error by a further $3$--$12\times$ over linear extrapolation on step-boundary models. Removing the language modeling loss yields trajectories that are $2\times$ more MLP-predictable than the combined loss, revealing a tradeoff between generation quality and geometric purity. Our results identify sampling position as the critical variable in geometric regularization and establish multi-step latent prediction MSE as a new evaluation metric for this class of methods. 
\end{abstract}

\section{Introduction}

Large Language models (LLMs) reason by generating step-by-step solutions, producing many tokens per reasoning step. Each token requires a full autoregressive forward pass, making multi-step reasoning computationally expensive. A growing body of work explores latent reasoning where computation is performed in the model's continuous hidden-state space rather than decoding discrete tokens at every step~~\citep{coconut,plat2026}. For latent reasoning to be feasible, the model's hidden-state trajectory must be predictable, which is given the current reasoning state, future states must be forecastable with reasonable accuracy.

Semantic Tube Prediction (STP)~\citep{stp2026} provides a theoretical and practical framework for this. Building on the Geodesic Hypothesis, which insists that error-free token sequences trace locally linear geodesics on a smooth semantic manifold, STP adds a supplementary cosine-displacement loss during fine-tuning that forces consecutive hidden-state displacement vectors to be parallel. The resulting ``Semantic Tube'' confines trajectories to a tubular neighborhood of the geodesic, improving signal-to-noise ratio and enabling $16\times$ data efficiency gains.

STP samples its loss at random token sub-spans, where three indices $s < r < t$ are drawn uniformly from the token sequence. While this method is elegant, we would like to explore if  the semantic structure of the reasoning process, which is the natural boundaries where the model transitions from one reasoning step to the next, can contribute to the geometric smoothness of the method. We hypothesize that aligning the geometric regularization with the semantic structure of reasoning at step boundaries rather than at arbitrary token positions will produce more predictable trajectories. We test whether step-boundary STP creates the latent-space conditions that future JEPA ~\citep{lecun2022} or COCONUT-style systems would need, as measured via post-hoc multi-step prediction accuracy of both linear and learned predictors.

\paragraph{Our key contributions.}
\begin{enumerate}
    \item \textbf{Sampling position is the critical variable.} Applying STP at consecutive step boundaries achieves $168\times$ prediction improvement vs.\ $4\times$ for random-token STP---a $40\times$ gap from where the loss is sampled, not what the loss computes (\S\ref{sec:main-result}).
    \item \textbf{Multi-step latent prediction MSE as evaluation metric.} We introduce and validate this metric with decoding fidelity tests: predicted embeddings decode to the correct next token 93.4\% of the time (\S\ref{sec:decoding}).
    \item \textbf{Trajectories are smooth curves, not straight lines.} A learned MLP predictor reduces error by $3$--$12\times$ over linear on STP-shaped models but finds no systematic structure on baselines, revealing three distinct manifold regimes (\S\ref{sec:skip-predictor}).
    \item \textbf{$\mathcal{L}_{\text{NTP}}$ creates a tradeoff.} Removing the language modeling loss yields $2\times$ better MLP prediction at the cost of 3.7\,pp GSM8K accuracy---generation quality vs.\ geometric purity (\S\ref{sec:main-result}, \S\ref{sec:skip-predictor}).
    \item \textbf{Negative finding.} Geometric smoothness does not encode step correctness (AUC $\approx 0.5$; \S\ref{sec:neg-correctness}): the Semantic Tube captures organized thinking, not correct thinking.
\end{enumerate}

\section{Background}

\subsection{Semantic Tube Prediction}

\citet{stp2026} propose the Geodesic Hypothesis, which describes that token sequences generated by LLMs trace geodesics on a smooth semantic manifold, and deviations from these geodesics represent noise. They formalize this as the STP loss:
\begin{equation}
    \mathcal{L}_{\text{STP}} = \mathbb{E}_{s < r < t}\left[1 - \cos(h_r - h_s,\; h_t - h_r)\right]
\end{equation}
where $h_s, h_r, h_t$ are hidden states at randomly sampled token positions. Minimizing $\mathcal{L}_{\text{STP}}$ forces consecutive displacement vectors to be parallel, confining trajectories to a tube around the geodesic. Combined with the standard next-token prediction loss, the training objective is $\mathcal{L} = \mathcal{L}_{\text{NTP}} + \lambda \cdot \mathcal{L}_{\text{STP}}$.

With the predictability of a geometric constraint, STP is able to dramatically reduce the amount of data required for training. Finetuning with $16\times$ less training data across multiple model families~\citep{stp2026} resulted in comparable results with full-data training.

\subsection{Joint Embedding Predictive Architectures}

We draw motivation from the JEPA paradigm~\citep{lecun2022}, which advocates predicting in latent space rather than reconstructing inputs. I-JEPA~\citep{ijepa} and V-JEPA~\citep{vjepa} demonstrate this for images and video; LLM-JEPA~\citep{llmjepa2025} extends it to language. Interestinly, STP was interestingly introduced to solve the two-view problem of LLM-JEPA, which requires the text data to be paired with the code form of the same information. We do not implement a JEPA architecture as our training lacks target encoder, learned predictor module, and continuous embedding feedback loop. Instead, we test the preconditions that the latent space supports accurate multi-step prediction.

\subsection{Geometric Analysis of LLM Hidden States}

Several works analyze the geometry of LLM hidden states diagnostically. \citet{wang2025} measure curvature across layers; \citet{zhou2026} apply Menger curvature to reasoning trajectories; \citet{herrmann2025} introduce PHi loss and show that hidden-state unpredictability correlates with correct solutions on difficult problems. \citet{traced2026} decompose reasoning traces into Progress (displacement magnitude) and Stability (curvature) at the token level. \citet{sun2026trajectories} extract activations at step markers and find that correct and incorrect solutions diverge geometrically at late reasoning steps. \citet{statphys2025} model reasoning trajectories as drift-diffusion systems with regime switching. \citet{damirchi2026} find that smooth layer-wise trajectories correlate with valid reasoning.

\subsection{Latent Reasoning}

Latent reasoning to reduce or avoid token-wise decoding has been explored in several recent works. COCONUT~\citep{coconut} feeds the last hidden state back as the next input embedding, enabling latent iteration without token decoding. PLaT~\citep{plat2026} decouples reasoning from verbalization entirely. CoLaR~\citep{colar2025v2} trains a latent head with a next-compressed-embedding objective at random token positions. LightThinker~\citep{lightthinker2025} compresses intermediate steps into gist tokens at step boundaries. \citet{nextlat2025} train a next-latent prediction head with Smooth L1 loss at every token, supporting multi-step rollouts for world modeling. At inference time, The Geometric Reasoner~\citep{geomreasoner2026} scores candidates via bumpiness penalties at chunk boundaries, and STEP~\citep{step2026} trains a classifier on step-boundary hidden states to detect reasoning errors.

\section{Method}

\subsection{STP at Semantic Step Boundaries}

\begin{figure}[t]
    \centering
    \includegraphics[width=0.9\linewidth]{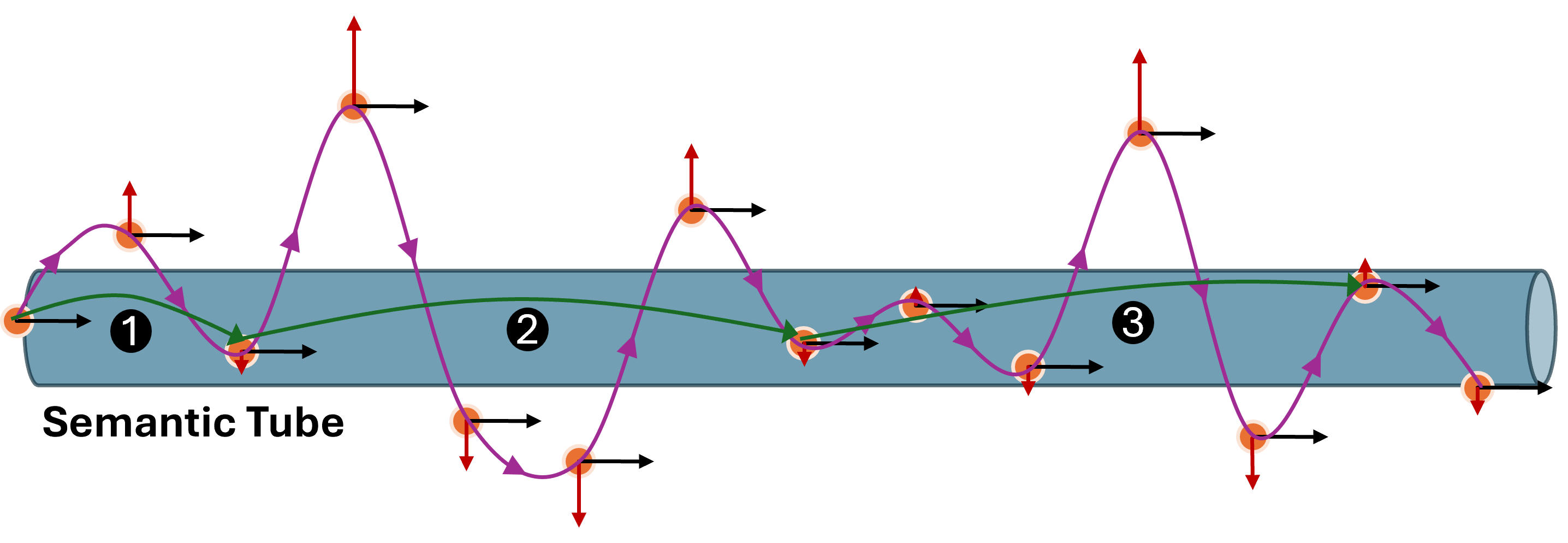}
    \caption{Semantic Step Prediction concept. The token-level trajectory (pink) oscillates around the step-level geodesic (green). Step boundary positions (numbered 1, 2, 3) define the trajectory $\mathbf{z} = (z_0, z_1, \ldots, z_K)$ on which the STP loss enforces consecutive displacement parallelism, confining the step-level path to a smooth tube.}
    \label{fig:concept}
\end{figure}

We insert a special \texttt{<|step|>} delimiter token between reasoning steps in the training data (Figure~\ref{fig:concept}), yielding sequences of the form:
\begin{equation}
    [\text{question}]\;\texttt{<|step|>}\;[\text{step}_1]\;\texttt{<|step|>}\;[\text{step}_2]\;\texttt{<|step|>}\;\cdots\;\texttt{<|step|>}\;[\text{step}_K]\;\texttt{<|step|>}
\end{equation}
The hidden states at \texttt{<|step|>} positions form a trajectory $\mathbf{z} = (z_0, z_1, \ldots, z_K)$. We compute the STP loss on consecutive triples:
\begin{equation}\label{eq:stp-step}
    \mathcal{L}_{\text{STP}}^{\text{step}} = \frac{1}{K-1} \sum_{k=1}^{K-1} \left(1 - \frac{(z_k - z_{k-1}) \cdot (z_{k+1} - z_k)}{\|z_k - z_{k-1}\| \cdot \|z_{k+1} - z_k\| + \epsilon}\right)
\end{equation}
This is mathematically identical to STP's loss but differs in where the indices are sampled. We sample at semantically meaningful step boundaries rather than at random token positions. The combined objective is $\mathcal{L} = \mathcal{L}_{\text{NTP}} + \beta \cdot \mathcal{L}_{\text{STP}}^{\text{step}}$ with $\beta = 1$.

Fine-tuning uses LoRA~\citep{lora} with rank 16 on $q, k, v, o$ projections ($\sim$4.4M trainable parameters on a 1.5B base model).

\subsection{Experimental Grid}\label{sec:grid}

\begin{table}[t]
    \caption{Six-model experimental grid. Each model differs in loss composition and/or STP sampling strategy.}
    \label{tab:grid}
    \centering
    \begin{tabular}{llll}
        \toprule
        model & $\mathcal{L}_{\text{NTP}}$ & $\mathcal{L}_{\text{STP}}$ & STP sampling \\
        \midrule
        B1  & ---            & ---            & --- (frozen pretrained) \\
        B2  & $\checkmark$   & ---            & --- (vanilla LM fine-tuning) \\
        C   & $\checkmark$   & $\checkmark$   & Random tokens (original STP) \\
        A2  & $\checkmark$   & $\checkmark$   & Random step boundaries \\
        \textbf{A}   & $\checkmark$   & $\checkmark$   & \textbf{Consecutive step boundaries} \\
        A1 & ---          & $\checkmark$   & Consecutive step boundaries \\
        \bottomrule
    \end{tabular}
\end{table}

This grid (Table~\ref{tab:grid}) isolates three factors: (i) token-level vs.\ step-level sampling (C vs.\ A2), (ii) random vs.\ consecutive at the step level (A2 vs.\ A), and (iii) necessity of $\mathcal{L}_{\text{NTP}}$ (A vs.\ A1).

\subsection{Multi-Step Prediction MSE}\label{sec:mse-def}

For each trajectory $(z_0, z_1, \ldots, z_K)$ and each valid position $k \geq 1$, we predict $m$ steps ahead via linear extrapolation:
\begin{equation}\label{eq:linear-pred}
    \hat{z}_{k+m} = z_k + m \cdot (z_k - z_{k-1})
\end{equation}
and compute the normalized prediction error:
\begin{equation}\label{eq:mse}
    \text{MSE}_m = \frac{1}{N} \sum_{(k,\text{sample})} \frac{\|\hat{z}_{k+m} - z_{k+m}\|^2}{\|z_{k+m}\|^2}
\end{equation}
$\text{MSE}_m \approx 0$ means the trajectory is perfectly linear; $\text{MSE}_m \approx 1$ means the prediction error equals the signal magnitude.

\subsection{Trajectory Smoothness Scores}

We report the cosine score (matching the training objective):
\begin{equation}
    \text{cos\_score}_k = 1 - \frac{(z_k - z_{k-1}) \cdot (z_{k+1} - z_k)}{\|z_k - z_{k-1}\| \cdot \|z_{k+1} - z_k\|} \in [0, 2]
\end{equation}
and the perpendicular score (geometric interpretation):
\begin{equation}
    \text{perp\_score}_k = \frac{\|d_k - (d_k \cdot \hat{s}_k)\hat{s}_k\|}{\|d_k\|} = \sin(\theta_k) \in [0, 1]
\end{equation}
where $d_k = z_k - z_{k-1}$ and $s_k = z_{k+1} - z_{k-1}$. The two scores are monotonically related and produce identical model orderings (Spearman $\rho = 1.0$).

\section{Experiments}

\subsection{Setup}

\textbf{models.} Qwen2.5-Math-1.5B (math-specialized), Qwen2.5-1.5B (general-purpose), Llama-3.2-1B (different architecture, 2048-dim hidden states vs.\ 1536). All fine-tuned via LoRA with rank 16.

\textbf{Training data.} 6,132 MATH competition problems and solutions split at paragraph boundaries with \texttt{<|step|>} delimiters. 3 epochs ($\sim$1,150 optimizer steps per model). 

\textbf{Evaluation data.} ProcessBench~\citep{processbench}: 3,400 step-by-step mathematical solutions across four difficulty levels with human-annotated error positions.

\section {Results}

\subsection{Prediction MSE}\label{sec:main-result}

\begin{table}[t]
    \caption{Six-model ablation on ProcessBench with Qwen2.5-Math-1.5B. Mean cosine score ($\downarrow$, matches training objective), mean perpendicular score ($\downarrow$), and multi-step prediction MSE ($\downarrow$). Improvement factor computed at $m{=}1$ relative to B1.}
    \label{tab:main}
    \centering
    \begin{tabular}{lrrrrr}
        \toprule
        model & Mean cos $\downarrow$ & Mean perp $\downarrow$ & MSE$_1$ $\downarrow$ & MSE$_3$ $\downarrow$ & $\times$ vs B1 \\
        \midrule
        B1 (frozen)       & 1.269 & 0.772 & 0.955 & 5.688 & 1$\times$ \\
        B2 (LM only)      & 1.288 & 0.822 & 0.600 & 2.797 & 1.6$\times$ \\
        C (rand-tok STP)   & 0.691 & 0.567 & 0.226 & 0.836 & 4$\times$ \\
        A2 (rand-step STP) & 0.018 & 0.075 & 0.010 & 0.155 & 96$\times$ \\
        \textbf{A (consec-step STP)} & \textbf{0.014} & \textbf{0.081} & \textbf{0.006} & \textbf{0.093} & \textbf{168$\times$} \\
        \textbf{A1 (STP only)}    & \textbf{0.005} & \textbf{0.047} & \textbf{0.006} & \textbf{0.116} & \textbf{152$\times$} \\
        \bottomrule
    \end{tabular}
\end{table}

\begin{figure}[t]
    \centering
    \includegraphics[width=0.85\linewidth]{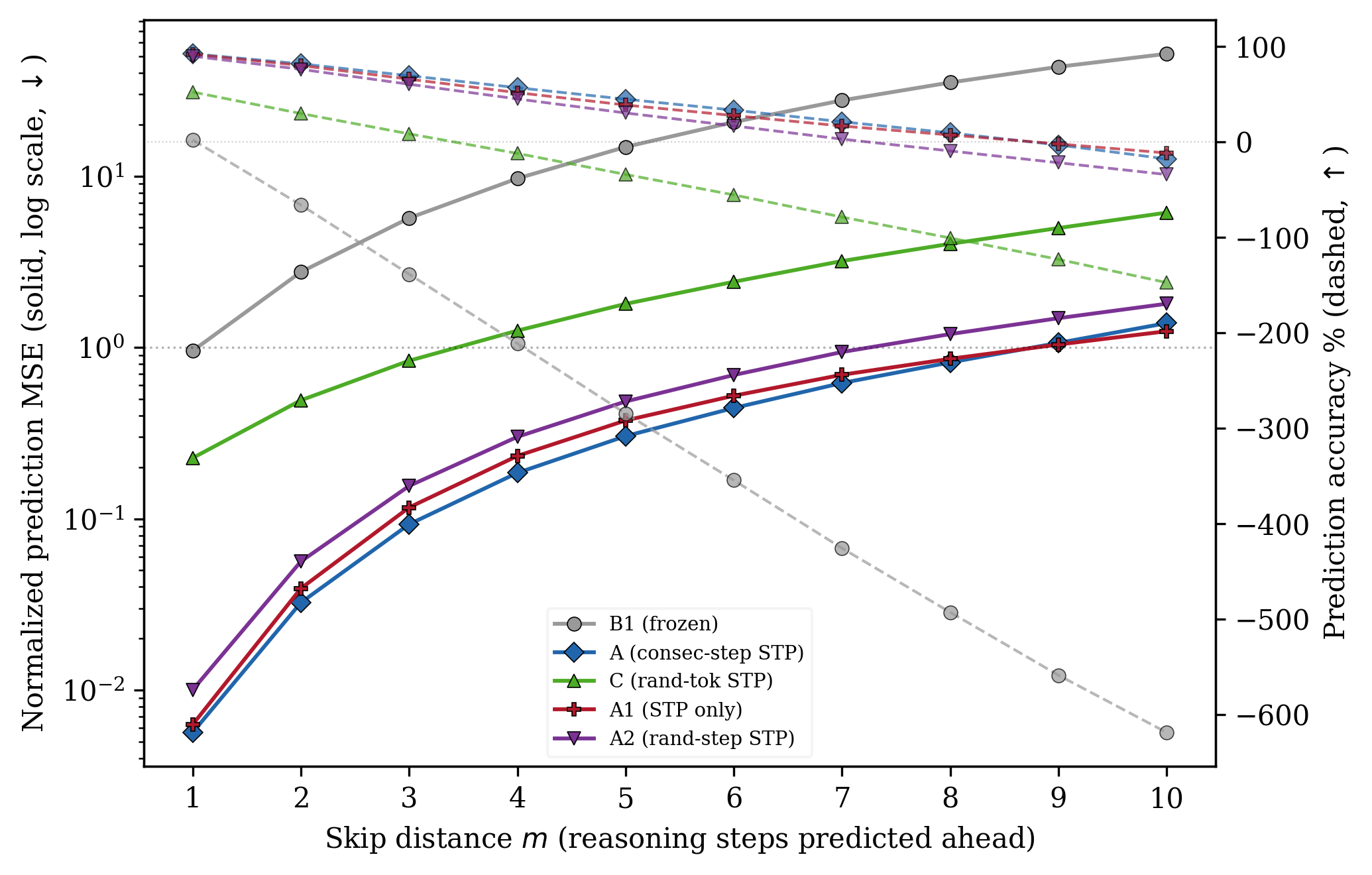}
    \caption{Multi-step prediction MSE (solid, left axis, log scale) and prediction accuracy (dashed, right axis) vs.\ skip distance $m$. Step-boundary models (A, A1, A2) achieve orders-of-magnitude better prediction than frozen (B1) and random-token STP (C). Model~A remains usable (MSE $< 1$) up to $m{=}8$.}
    \label{fig:mse-skip}
\end{figure}

Sampling at consecutive step boundaries (model~A) achieves $168\times$ more accurate prediction than the frozen baseline at $m{=}1$, while the original random-token recipe (model~C) achieves only $4\times$ (Table~\ref{tab:main}, Figure~\ref{fig:mse-skip}). Model~A's prediction at $m{=}3$ (MSE 0.093) is more accurate than B1's prediction at $m{=}1$ (MSE 0.955). The role of $\mathcal{L}_{\text{NTP}}$ sees to be a tradeoff and not a redundancy. Model A1 ($\mathcal{L}_{\text{STP}}^{\text{step}}$ only) tests whether $\mathcal{L}_{\text{NTP}}$ can be dropped. Under linear extrapolation, A1 matches A within 12\%. Under a learned MLP predictor, A1 outperforms A by 47--57\% across all skip distances (\S\ref{sec:skip-predictor}). The choice is a tradeoff between generation quality ($\mathcal{L}_{\text{NTP}}$: +3.7\% GSM8K) and geometric predictability where $\mathcal{L}_{\text{STP}}$ alone: $2\times$ better MLP prediction.

\subsection{Cross-model and Cross-Dataset Validation}

\begin{table}[t]
    \caption{B1 vs.\ A across three model families on ProcessBench. Accuracy $= 1 - \sqrt{\text{MSE}}$.}
    \label{tab:cross-model}
    \centering
    \begin{tabular}{lrrrr}
        \toprule
        model & B1 MSE$_1$ & A MSE$_1$ & Improvement & A acc.\ $m{=}1$ \\
        \midrule
        Qwen2.5-Math-1.5B    & 0.955 & 0.006 & 168$\times$ & 92.5\% \\
        Llama-3.2-1B          & 0.503 & 0.002 & 335$\times$ & 96.1\% \\
        Qwen2.5-1.5B (general)& 0.678 & 0.018 & 39$\times$  & 86.8\% \\
        \bottomrule
    \end{tabular}
\end{table}

\begin{figure}[t]
    \centering
    \includegraphics[width=\linewidth]{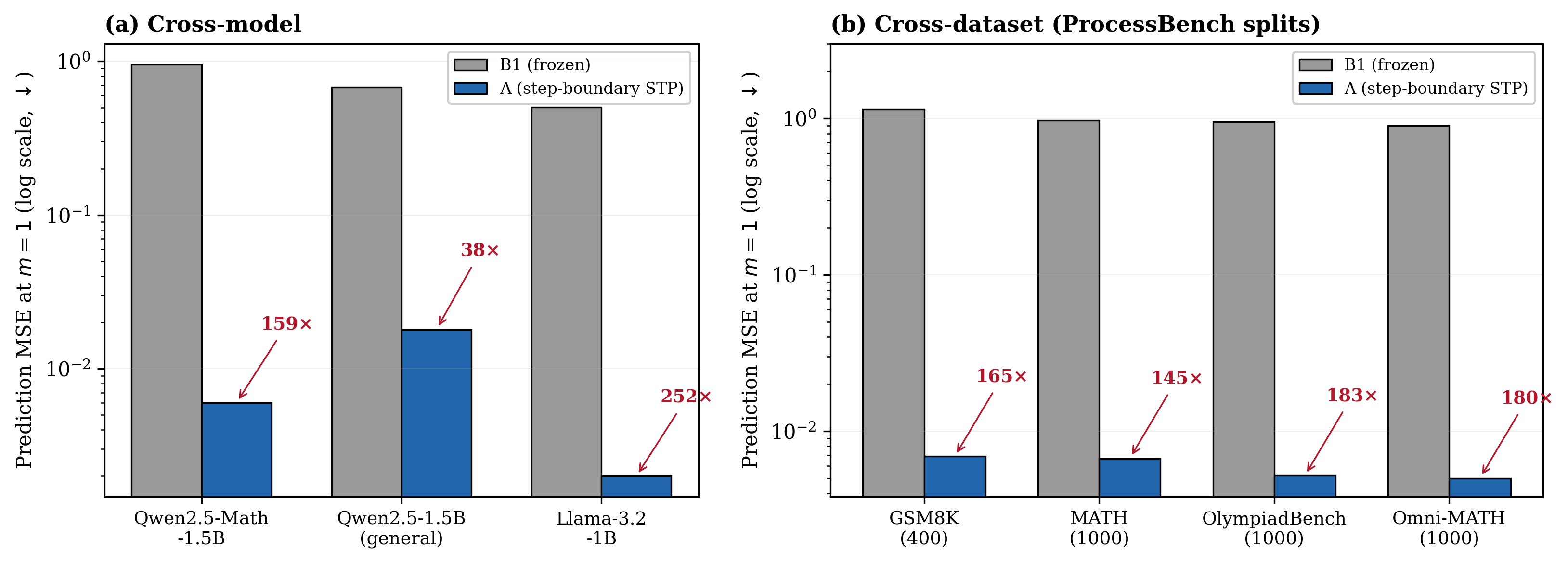}
    \caption{Robustness of the geometric improvement. \textbf{(a)}~Cross-model: B1 vs.\ A prediction MSE at $m{=}1$ across three model families (log scale). The improvement ranges from $39\times$ (Qwen2.5-1.5B) to $335\times$ (Llama-3.2-1B). \textbf{(b)}~Cross-dataset: B1 vs.\ A on four ProcessBench splits of increasing difficulty. The improvement is consistent ($145\times$--$183\times$) and slightly stronger on harder datasets.}
    \label{fig:robustness}
\end{figure}

To investigate if geometric representation finetuning is transferable to non-math models, we similarly finetuned two models: Qwen2.5-1.5B, which is a direct comparison to Qwen2.5-Math-1.5B, as well as Llama-3.2-1B, which represents a different model architecture with different dimensions. From the results, we observed that the effect transfers across architectures, hidden dimensions, and pretraining regimes (Table~\ref{tab:cross-model}, Figure~\ref{fig:robustness}a). Llama-3.2-1B achieves the strongest improvement ($335\times$) despite having no math-specific pretraining.

The improvement also holds across difficulty levels (Figure~\ref{fig:robustness}b) on four ProcessBench splits, GSM8K ($165\times$), MATH ($145\times$), OlympiadBench ($183\times$), and Omni-MATH ($180\times$). The geometric shaping is consistent and slightly stronger on harder problems, thereby ruling out the concern that STP overfits to easy problem structure.

\subsection{Task Accuracy: No Degradation}

\begin{table}[t]
    \caption{GSM8K accuracy (greedy decoding).}
    \label{tab:gsm8k}
    \centering
    \begin{tabular}{lrrrl}
        \toprule
        model & Accuracy & $n$ & MSE$_1$ & Interpretation \\
        \midrule
        C (rand-tok STP)    & 66.9\% & 882/1319  & 0.226 & Hurts accuracy \\
        B1 (frozen)         & 68.8\% & 907/1319  & 0.955 & Pretrained baseline \\
        A1 (STP only)    & 69.3\% & 914/1319  & 0.006 & No $\mathcal{L}_{\text{NTP}}$: preserves acc. \\
        \textbf{B2 (LM only)}& \textbf{73.0\%} & \textbf{963/1319} & \textbf{0.600} & Vanilla fine-tuning \\
        \textbf{A (consec-step)} & \textbf{73.0\%} & \textbf{963/1319} & \textbf{0.006} & Matches B2, 168$\times$ better MSE \\
        \bottomrule
    \end{tabular}
\end{table}

Overall, all models perform somewhat equally on accuracy, confirming that step-boundary STP does not degrade the model's reasoning ability. Model A achieves accuracy identical to vanilla fine-tuning (B2) while simultaneously achieving $168\times$ better latent prediction MSE (Table~\ref{tab:gsm8k}). Random-token STP is the only variant that showed a decrease in accuracy relative to the frozen baseline. This suggests that the semantic structure of reasoning prevents interference with the language model objective. The accuracy and MSE tests in fact reveal that two losses contribute orthogonally. $\mathcal{L}{\text{NTP}}$ seem to impact accuracy improvement while $\mathcal{L}{\text{STP}}$ brings in geometric alignment to shape how models think through latent space.

\subsection{Decoding Validation}\label{sec:decoding}

\begin{table}[t]
    \caption{Decoding from predicted embeddings (200 ProcessBench samples).}
    \label{tab:decoding}
    \centering
    \begin{tabular}{llrr}
        \toprule
        Metric & $m$ & model A & B1 (frozen) \\
        \midrule
        \multirow{3}{*}{Step retrieval acc.\ $\uparrow$} & 1 & \textbf{92.2\%} & 70.2\% \\
        & 2 & \textbf{55.7\%} & 14.9\% \\
        & 3 & \textbf{47.5\%} & 10.1\% \\
        \midrule
        \multirow{3}{*}{Top-1 token agreement $\uparrow$} & 1 & \textbf{93.4\%} & 19.1\% \\
        & 2 & \textbf{71.1\%} & 10.1\% \\
        & 3 & \textbf{34.1\%} & 8.3\% \\
        \midrule
        Top-5 Jaccard $\uparrow$ & 1 & \textbf{0.878} & 0.207 \\
        KL divergence $\downarrow$ & 1 & \textbf{0.019} & 4.581 \\
        \bottomrule
    \end{tabular}
\end{table}

The decoding validation (Table~\ref{tab:decoding}) experiment shows that two hidden-state vectors can be close in Euclidean distance yet differ in the specific subspace that the LM head uses for next-token prediction. The information relevant to decoding occupies a lower-dimensional manifold within the full 1536-dimensional hidden-state space. To test whether our predicted embeddings are functionally equivalent and not merely geometrically close, we pass both the predicted $\hat{z}_{k+m}$ and the actual $z_{k+m}$ through the model's final RMSNorm and LM head, producing two probability distributions over the vocabulary. At $m{=}1$, the distributions agree on the top-1 token 93.4\% of the time (the frozen baseline B1 had 19.1\%), and with a KL divergence of just 0.019 nats (vs.\ 4.581 for B1). The step retrieval metric provides a complementary view: the predicted embedding's nearest neighbor among all step-boundary embeddings (excluding the source points $z_k$ and $z_{k-1}$) is the correct target step 92.2\% of the time, confirming that $\hat{z}_{k+m}$ lands at the right reasoning state. An instructive contrast emerges between model~A and B1 at $m{=}1$: B1 achieves relatively high step retrieval (70.2\%) but low token agreement (19.1\%), meaning the frozen model's predicted embedding is geometrically near the correct step but encodes different next-token information---the geometric position and functional content are misaligned. STP training aligns both: model~A's 92.2\% retrieval and 93.4\% token agreement show that geometric proximity and functional equivalence coincide after geometric shaping. At longer horizons, model~A's token agreement degrades gracefully---71.1\% at $m{=}2$ and 34.1\% at $m{=}3$---while B1 collapses to near-chance (10.1\% and 8.3\%). Notably, model~A at $m{=}2$ (71.1\%) still exceeds B1 at $m{=}1$ (19.1\%), mirroring the MSE result and confirming that the prediction advantage extends well beyond one-step forecasting.

\section{Analysis}

\subsection{Manifold Structure}\label{sec:skip-predictor}

\begin{table}[t]
    \caption{MLP vs.\ linear extrapolation on ProcessBench (all six models). Ratio = MLP MSE / Linear MSE; decreasing ratio with $m$ indicates smooth curvature.}
    \label{tab:skip}
    \centering
    \small
    \begin{tabular}{lrrrrl}
        \toprule
        model & $m$ & Linear MSE & MLP MSE & Ratio & Trend \\
        \midrule
        \multirow{3}{*}{B1 (frozen)} & 1 & 0.954 & 0.385 & 0.403 & \multirow{3}{*}{weak $\downarrow$} \\
        & 2 & 2.770 & 0.975 & 0.352 & \\
        & 3 & 5.732 & 2.004 & 0.350 & \\
        \midrule
        \multirow{3}{*}{B2 (LM only)} & 1 & 0.594 & 0.270 & 0.454 & \multirow{3}{*}{flat} \\
        & 2 & 1.482 & 0.635 & 0.428 & \\
        & 3 & 2.785 & 1.214 & 0.436 & \\
        \midrule
        \multirow{3}{*}{C (rand-tok STP)} & 1 & 0.225 & 0.105 & 0.466 & \multirow{3}{*}{moderate $\downarrow$} \\
        & 2 & 0.488 & 0.173 & 0.354 & \\
        & 3 & 0.825 & 0.266 & 0.322 & \\
        \midrule
        \multirow{3}{*}{A2 (rand-step)} & 1 & 0.010 & 0.00235 & 0.233 & \multirow{3}{*}{strong $\downarrow$} \\
        & 2 & 0.056 & 0.00554 & 0.099 & \\
        & 3 & 0.155 & 0.00952 & 0.061 & \\
        \midrule
        \multirow{3}{*}{A (consec-step)} & 1 & 0.00568 & 0.00162 & 0.285 & \multirow{3}{*}{strong $\downarrow$} \\
        & 2 & 0.03223 & 0.00403 & 0.125 & \\
        & 3 & 0.09240 & 0.00785 & 0.085 & \\
        \midrule
        \multirow{3}{*}{\textbf{A1 (STP only)}} & 1 & \textbf{0.00635} & \textbf{0.000868} & \textbf{0.137} & \multirow{3}{*}{\textbf{strongest $\downarrow$}} \\
        & 2 & 0.03952 & 0.002141 & 0.054 & \\
        & 3 & 0.11722 & 0.003407 & 0.029 & \\
        \bottomrule
    \end{tabular}
\end{table}

We train a residual multilayer perceptron (MLP) post-hoc predictor probe, $\hat{z}_{k+m} = z_k + m(z_k - z_{k-1}) + g_\phi(z_k, z_{k-1})$, where $g_\phi$ is a 3-layer MLP with zero-initialized last layer (architecture details in Appendix~\ref{app:skip-predictor}). The zero initialization ensures that the MLP starts as an exact copy of the linear predictor and can only improve upon it. Any reduction in MSE represents a genuine non-linear structure that linear extrapolation missed. The MLP probe is trained on 80\% of ProcessBench step-boundary pairs and evaluated on the held-out 20\%. The probe serves purely as a measurement instrument for the latent geometry of the finetuned model without changing the models.  

\begin{figure}[t]
    \centering
    \includegraphics[width=0.65\linewidth]{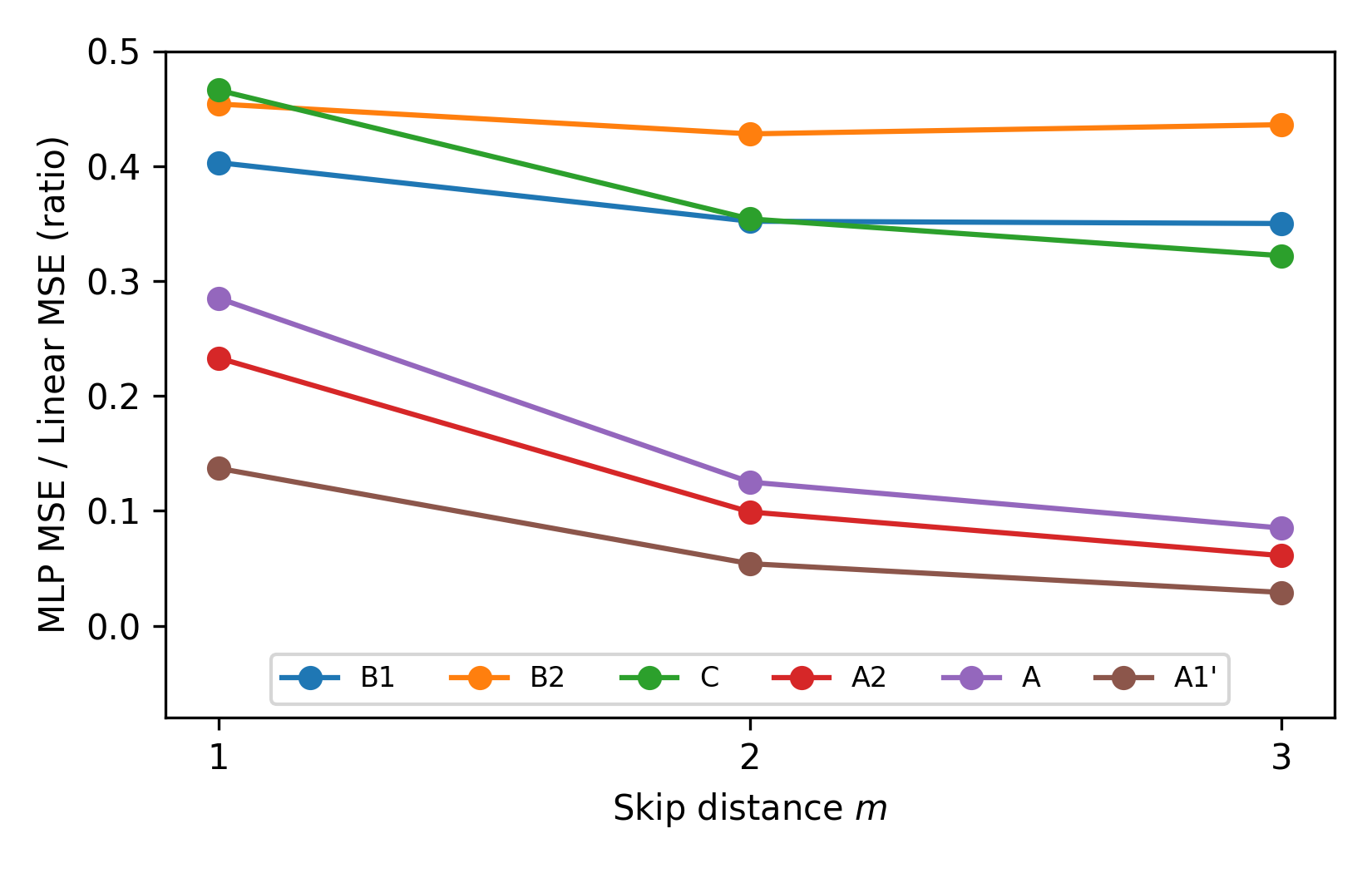}
    \caption{MLP/linear MSE ratio vs.\ skip distance $m$ for all six models. Flat ratio (B1, B2, C) indicates noise without systematic curvature. Decreasing ratio (A2, A, A1) is the signature of a smoothly curving manifold where curvature compounds with $m$. A1 (STP only) has the steepest decrease.}
    \label{fig:ratio}
\end{figure}

To characterize model's manifold structure, we compared the MLP against a trivial linear probe on the same task of predicting the hidden state at step $k+m$ when given hidden state of step $k$ and step $k-1$. The ratio of MLP MSE to and liner MSE reveals how much learnable strcture exists beyond what a striaght-line approximation captures. Crucially, a ratio that decreases with $m$ indicates smooth, compounding carvature, which is the signature of a coherent curving manifold rather than random noise. (Table~\ref{tab:skip}, Figure~\ref{fig:ratio})

Models B1, B2 and C show flat or weakly decreasing ratios, while models with step-boundary finetuning, such as A, A1, and A2, show strongly decreasing ratios. Model A goes from 0.285 to 0.085 (a 70\% drop over  $m{=}1$ to $m{=}3$), and A1$'$ goes from 0.137 to 0.029 (a 79\% drop). Decreasing ratio with $m$ is the signature of a smoothly curving path. curvature compounds quadratically with horizon, so longer extrapolations expose proportionally more non-linear structure for the MLP to learn. A1 (pure $\mathcal{L}{\text{STP}}$, no language modeling loss) has the steepest decrease and lowest ratios of any cell, producing the cleanest geometric structure. It's MLP at $m{=}3$ achieves MSE 0.003407 --- a $1{,}098\times$ improvement over B1's linear prediction at $m{=}1$ (MSE 0.954). With a learned predictor and pure geometric training, we can forecast three reasoning steps ahead more accurately than a frozen pretrained model can forecast one.

\subsection{Why Sampling Position Matters}\label{sec:theory}

The $40\times$ gap between model~A and model~C arises from optimization efficiency: step-level STP concentrates all gradient signal on the $K$ step-boundary positions that define the trajectory we measure, while token-level STP distributes $q$ triples across $T$ token positions with $O(1/n)$ dilution per cross-step triple. The effective gradient ratio is $\sim n(K{-}1)/q \approx 100\times$ for typical values. Both models use identical LoRA parameters, training data, epochs, and compute. Hence, likely the $40\times$ MSE gap is entirely attributable to sampling position.

The MLP/linear ratio decreasing with $m$ (Table~\ref{tab:skip}) follows from curvature bounds. The linear prediction error scales as $O(m^2\epsilon)$ while a curved predictor achieves $O(m^2\epsilon^2)$, yielding ratio $O(\epsilon)$. Model~A1 ($\epsilon = 0.005$) has lower ratios than model~A ($\epsilon = 0.014$) at every $m$, consistent with the $O(\epsilon)$ prediction. The $2\times$ MLP gap between A1 and A follows from gradient interference: $\mathcal{L}_{\text{NTP}}$ adds perturbations orthogonal to the geometric direction that linear tolerates but the MLP cannot fit through.

Our results establish a clear hierarchy of latent-space support for multi-step prediction. Frozen pretrained representations (model~B1) that are based on token-based sampling are not suitable for prediction. The linear MSE $\approx 1.0$ at $m{=}1$, means the prediction error equals the signal magnitude, faired worse compared to a simple learned MLP ( reduced to MSE $\approx 0.39$). This confirmed that the frozen trajectory has no coherent geometric structure for a neural predictor to exploit. Vanilla LM fine-tuning (model~B2) reduces linear MSE to 0.60, but the MLP/linear ratio remained flat at $\sim$0.45 across all skip distances. Random-token STP (model~C) provided a further reduction to linear MSE at $\approx 0.23$, however, its MLP/linear ratio (0.466) is the highest of any cell, suggesting the residual structure is less learnable than the frozen baseline. This may mean that Random-token STP denoises but it does not create a manifold. Only step-boundary STP showed capability in predictable trajectories. Model~A achieved linear MSE of $\approx 0.006$ with a strongly decreasing MLP/linear ratio (0.285 $\to$ 0.085). The gap between Cell~C (MSE 0.226) and Cell~A (MSE 0.006) is $40\times$, arising entirely from where the loss is sampled, not from what the loss computes. Model~A1, without the $\mathcal{L}{\text{NTP}}$ pushed further to MLP MSE $\approx 0.0009$, about a $1{,}098\times$ improvement over B1's linear baseline.

\section{Limitations and Future Work}

\textbf{Limitations.} (1)~We characterize prediction potential but do not build a latent reasoning system; utilizing the geometric structure for generation would require COCONUT-style architectural changes~\citep{coconut}. (2)~Step boundaries are manually marked via \texttt{<|step|>} at paragraph breaks; some MATH solutions have noisy boundaries (e.g., individual lines of a \LaTeX{} \texttt{align*} block). Despite this, the $168\times$ improvement is robust. (3)~Evaluation is on math reasoning only. (4)~MSE at $m{=}1$ is closely related to the STP training objective; MSE at $m \geq 2$ and the MLP predictor provide independent validation.

\textbf{Future work.} (1)~Combining step-boundary STP with COCONUT curriculum training---STP straightens the trajectory that COCONUT iterates along. (2)~Data efficiency evaluation. (3)~Automatic step boundary detection. (4)~Cross-domain evaluation.

\section{Conclusion}

In this paper, we investigates the effect of STP's geometric loss and step sampling on shaping language model's reasoning trajectories. We showed that changing STP from random token sub-spans to semantic step boundaries transforms hidden-state trajectories from unpredictable walks (linear MSE $\approx 1.0$) into smoothly curving manifolds (linear MSE $\approx 0.006$; MLP MSE $\approx 0.0009$). And also achieved multi-step latent prediction MSE $168\times$ to $1{,}098\times$ improvement validated across three model families and four datasets. 

\section*{Code Availability}
The code used to reproduce all experiments and figures in this work is publicly available at \url{hhttps://github.com/YYDreamzure/SSP/}. The repository includes training scripts, evaluation pipelines, and pre-trained model checkpoints. Any additional data or materials required to reproduce the results reported in this paper are available from the corresponding author upon reasonable request.

{\small
\bibliographystyle{plainnat}

}

\appendix

\section{Supplementary Material}

\subsection{Perpendicular Score Geometry}

\begin{figure}[h]
    \centering
    \includegraphics[width=0.7\linewidth]{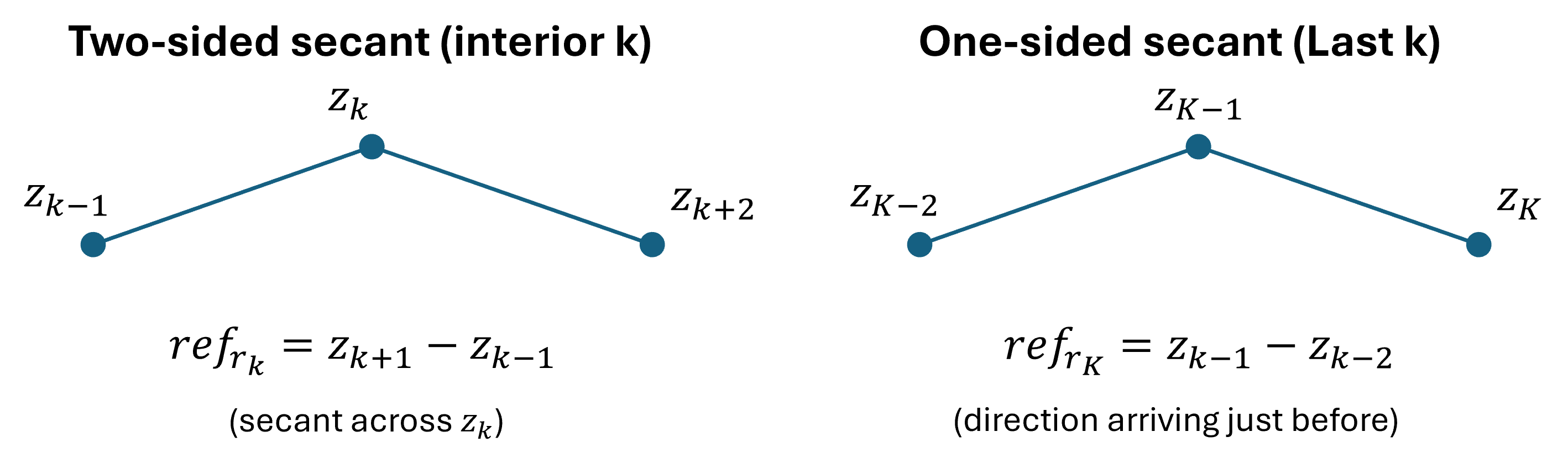}
    \caption{Perpendicular score computation. \textbf{Left}: interior positions use the two-sided secant $\text{ref}_{r_k} = z_{k+1} - z_{k-1}$ as the reference direction. \textbf{Right}: the last position uses the one-sided secant $\text{ref}_{r_K} = z_{K-1} - z_{K-2}$ (the arriving direction). The perpendicular score is $\sin\theta_k \in [0,1]$ where $\theta_k$ is the angle between the displacement $d_k$ and its projection onto the reference direction.}
    \label{fig:legvssecant}
\end{figure}

Binary detection AUC is indistinguishable from random (Table~\ref{tab:correctness}). The Semantic Tube captures the \emph{process} of organized reasoning---a structured, predictable flow through latent space---not the \emph{outcome} of correct reasoning. This is consistent with the PHi loss literature~\citep{herrmann2025}: hidden-state unpredictability correlates with correct solutions on difficult problems.

This finding resolves an apparent conflict with \citet{damirchi2026}, who report that smooth trajectories correlate with valid reasoning. The two analyses operate on different axes: smoothness across \emph{layers} (vertical) may reflect well-conditioned computation, while smoothness across \emph{steps} (horizontal) reflects predictable reasoning content---which need not correlate with correctness.

\subsection{Smoothness $\neq$ Correctness}\label{sec:neg-correctness}

\begin{table}[t]
    \caption{ProcessBench error detection. Perpendicular score does not encode correctness.}
    \label{tab:correctness}
    \centering
    \begin{tabular}{lrrr}
        \toprule
        model & Binary AUC & Localization acc. & Random baseline \\
        \midrule
        B1 (frozen) & 0.509 & 6.6\% & $\sim$15--20\% \\
        A (step-STP) & 0.564 & 4.8\% & $\sim$15--20\% \\
        \bottomrule
    \end{tabular}
\end{table}

\subsection{Skip-Predictor Architectures}\label{app:skip-predictor}

We use two post-hoc predictors to probe the geometric structure of step-boundary trajectories. Both take as input the hidden states at positions $k$ and $k{-}1$ and predict the hidden state at position $k{+}m$, without modifying the underlying model.

\paragraph{Linear predictor (zero parameters).} Extrapolates along the current displacement direction:
\begin{equation}
    \hat{z}_{k+m} = z_k + m \cdot (z_k - z_{k-1})
\end{equation}
This assumes the trajectory is locally linear. If consecutive displacements are perfectly parallel ($\mathcal{L}_{\text{STP}}^{\text{step}} = 0$), linear prediction is exact. Any prediction error reflects curvature or noise in the trajectory.

\paragraph{MLP predictor (learned, $\sim$6M parameters).} Adds a learned non-linear correction on top of the linear baseline:
\begin{equation}
    \hat{z}_{k+m} = \underbrace{z_k + m \cdot (z_k - z_{k-1})}_{\text{linear baseline}} + \underbrace{g_\phi(z_k,\; z_{k-1})}_{\text{learned correction}}
\end{equation}
where $g_\phi$ is a 3-layer MLP (input: $2D \to 2048 \to 2048 \to D$, GELU activations). The last layer is zero-initialized, so at epoch~0 the MLP prediction equals the linear prediction exactly---it can only improve, never hurt. Training minimizes MSE on 80\% of ProcessBench step-boundary pairs; evaluation uses the held-out 20\%.

The ratio $\text{MSE}_m^{\text{MLP}} / \text{MSE}_m^{\text{linear}}$ characterizes the manifold structure: ratio $\approx 1$ means the trajectory is maximally linear (MLP adds nothing); ratio $\ll 1$ means systematic non-linear structure exists that the MLP exploits. A ratio that \emph{decreases} with $m$ indicates smooth curvature that compounds over longer horizons---the signature of a curving tube rather than a straight line or random walk.
\end{document}